\documentclass{article}

\PassOptionsToPackage{numbers, compress}{natbib}

\usepackage[preprint]{neurips_2026}
\usepackage{graphicx}
\usepackage[utf8]{inputenc} 
\usepackage[T1]{fontenc}    
\usepackage{hyperref}       
\usepackage{url}            
\usepackage{booktabs}       
\usepackage{amsfonts}       
\usepackage{nicefrac}       
\usepackage{microtype}      
\usepackage{xcolor}         
\usepackage{amsmath}

\title{Hallucination-Free GUI Grounding via Regression-Free Layout-Aware Matching}

\author{%
  Yuke Li\textsuperscript{1} \qquad
  Xuehan Hou\textsuperscript{1} \qquad
  Jie Chen\textsuperscript{1}$^{\dagger}$ \\
  \textsuperscript{1}School of Electronic and
  Computer Engineering, \\
  Peking University, \\
  Shenzhen, China \\
  \texttt{2301212734@stu.pku.edu.cn},
  \texttt{2401212849@stu.pku.edu.cn} \\
  {\small $^{\dagger}$\,Corresponding author: \texttt{jiechen2019@pku.edu.cn}}
}

\begin{document}

\maketitle

\begin{abstract}
GUI agents are shifting from metadata-dependent large language models to purely visual multimodal large language models (MLLMs) that operate directly on screenshots. The core task, GUI grounding, requires translating abstract user instructions into precise element coordinates. This task faces a persistent dual obstacle: conventional grounding models lack the semantic richness to interpret abstract instructions, while end-to-end MLLMs suffer from coordinate hallucinations caused by deficient fine-grained perception. We propose a regression-free framework where a frozen MLLM performs instruction parsing and a dedicated grounding model handles precise localization without learning any coordinate regression. A frozen MLLM first elaborates the abstract instruction into a structured visual description rich in layout cues. These descriptions are then fed to a novel Layout-Aware GUI Grounding Model, which performs regression-free localization by matching against layout-prior candidates, inherently suppressing hallucinations and avoiding expensive fine-tuning. The grounding model is trained with only Text/Icon binary labels, requiring no coordinate regression parameters. On ScreenSpot-Pro, our method achieves over 20\% improvement in grounding accuracy over end-to-end systems; on Mind2Web, it raises success rate and element selection rate by more than 15\%. These results demonstrate that decoupling instruction understanding from layout-aware localization effectively resolves the core challenges of GUI interaction.
\end{abstract}

\section{Introduction}

Teaching GUI agents to execute instructions such as ``close all windows'' demands substantially more than recognizing buttons on a screen. It requires the agent to read interface text, comprehend spatial relationships among elements, and map ambiguous natural language to precise pixel coordinates---all without access to the underlying application metadata that developers ordinarily take for granted. Consider a typical multi-window desktop scenario: File Explorer, a browser, and a document editor each possess their own title bar and a small close button in the upper‑right corner. A human instinctively associates the title ``File Explorer'' with the corresponding close icon, whereas an agent that confuses titles or misjudges the spatial extent of ``upper‑right'' will close the wrong window, or fail entirely. Endowing GUI agents with this seemingly ordinary human capability---robust, text‑aware, layout‑sensitive localization---is the central problem this paper sets out to solve.

Existing approaches each encounter fundamental bottlenecks. Methods that parse interface metadata (e.g., HTML DOM trees or Android view hierarchies) are often inapplicable in sandboxed or secure environments where such structural data is unavailable~\cite{kim,zheng,zhou,wu2024mobilevlm,wang2023enabling}. To circumvent metadata dependence, end‑to‑end multimodal large language models (MLLMs) such as CogAgent and SeeClick directly predict coordinates from screenshots~\cite{cogagent,zhang2023you,fuyu}. However, their visual encoders, designed for natural images, lack inductive biases for GUI layout regularities, leading to persistent coordinate hallucinations — geometrically plausible yet incorrect predictions — and failure on spatially qualified instructions~\cite{seeclick,omniparser,zheng2024gpt,aria,gou2024navigating}. Engineering wrappers do not resolve this: OpenCLaw, a highly engineered GUI agent, still relies on the underlying MLLM's zero‑shot visual capacity. Traditional grounding models such as Grounding DINO~\cite{groundingdino,ren2024grounding} further suffer from weak text understanding and no spatial layout perception, making them helpless for text‑spatial queries. Fine‑tuning on coordinate supervision can partially suppress hallucinations but demands expensive manual annotations and generalizes poorly. These three paths converge on a single hard truth: the GUI grounding field lacks a model that natively understands interface layout structure and supports open‑vocabulary matching without costly coordinate regression or external markup. Building that missing model is our goal.
\begin{figure}[t]
\begin{center}
\includegraphics[width=1.0\linewidth]{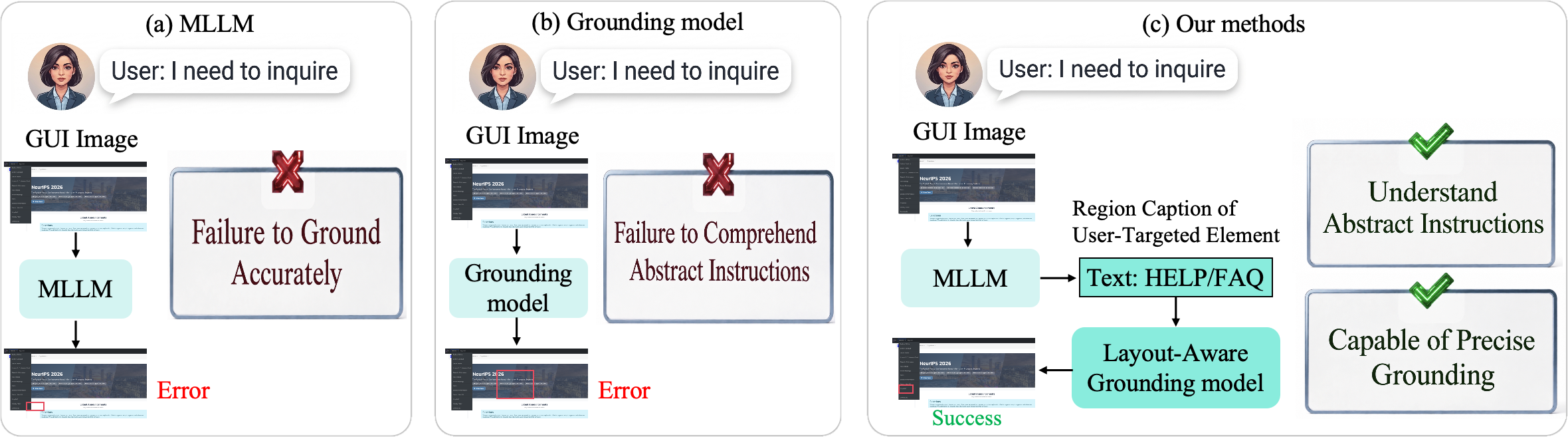}

\end{center}
   \caption{Framework of our proposed method. This approach integrates a Multimodal Large Language Model (MLLM) to interpret user instructions and produce detailed visual caption. These descriptions are subsequently utilized by a specially designed layout-aware grounding model to pinpoint the locations of user interface (UI) elements with enhanced precision.}
\label{fig:overview}
\end{figure}

Designing such a model means satisfying three constraints simultaneously: it must predict precise pixel coordinates without ever learning a coordinate regression mapping; it must internalize the basic layout primitives of a GUI---text blocks and icons---without requiring instance‑level semantic labels; and it must handle spatially modified queries (``upper‑right,'' ``left side'') without per‑query spatial annotations, while remaining robust when spatial information is irrelevant. These constraints give rise to three tightly related difficulties. First, frozen cross‑modal matching risks losing spatial precision, yet any reintroduction of regression opens the door to hallucinations; the pipeline must remain regression‑free while still producing fine‑grained, coordinate‑level accuracy. Second, standard detectors do not distinguish text from icons and treat GUIs as indistinct visual patches, but full category‑level bounding‑box annotation is prohibitively expensive; the model must acquire this discriminative layout sense from only a handful of images labeled with two classes---Text and Icon. Third, spatial queries require positional bias, yet brute‑force annotation of every possible spatial relation does not scale, and a heavy positional encoding head easily overfits while harming performance on spatially neutral queries; spatial awareness must be injected through an extremely lightweight mechanism that learns to shift attention when spatial cues are present and to remain nearly silent otherwise.

We address all three difficulties together with a unified Layout‑Aware GUI Grounding Model, built upon a decoupled two‑stage architecture. A fully frozen MLLM first converts the user instruction and screenshot into a structured textual description rich in layout cues---for example, ``the close button in the upper‑right corner of the `File Explorer' window''---without producing any coordinates. The core of our model, a dedicated grounding module, then takes this description and localizes the target through two tightly integrated steps: it generates a compact set of layout‑prior candidates by domain‑adapting a detector on a small set of GUI images labeled only with Text and Icon, and it selects the best‑matching candidate via frozen CLIP cross‑modal similarity, entirely without learning any coordinate regression parameters. A lightweight geometric fusion module---a single learnable linear projection that injects spatial encoding into the visual feature---handles spatially sensitive queries with negligible overhead, while remaining nearly silent for plain queries. In our running example, the ``File Explorer'' text candidate naturally scores highly against the generated description, and the geometric fusion nudges the highest score toward the actual close icon, delivering a clean, regression‑free hit.

Every component of this design directly targets a specific weakness of existing approaches. The frozen matching pathway eliminates the regression‑induced hallucinations that plague MLLMs. Layout‑prior candidate generation compensates for the layout blindness of generic detectors and end‑to‑end models. Lightweight geometric fusion handles spatially modified queries without demanding costly spatial annotations. Importantly, the entire grounding module is trainable with only a small amount of generic layout labels and a handful of spatial training pairs. On ScreenSpot-Pro, our method achieves over 20\% improvement in grounding accuracy over end‑to‑end state‑of‑the‑art systems, and on Mind2Web it raises the success rate and element selection rate by more than 15\%. Controlled ablation studies further confirm that layout‑prior candidate generation and lightweight geometric fusion each contribute substantially to these gains.

This work makes the following contributions in overview:
\begin{itemize}
    \item \textbf{Propose a dedicated Layout‑Aware GUI Grounding Model.} We introduce a visual grounding model specifically designed for interface layout structures and open‑vocabulary matching. By replacing coordinate regression with frozen cross‑modal matching, this model fundamentally prevents coordinate hallucinations while delivering precise pixel‑level localization.
    \item \textbf{Design a layout‑prior injection and lightweight geometric fusion mechanism.} This mechanism, using only a small set of Text/Icon‑level annotations, endows the grounding model with both text semantic understanding and spatial layout perception. It eliminates the dependency on expensive fine‑grained coordinate supervision and enables the model to handle spatially sensitive queries with negligible parameter overhead.
    \item \textbf{Build a decoupled intent‑driven grounding framework.} We embed the proposed grounding model into a two‑stage framework where a frozen MLLM handles instruction parsing and our model performs precise localization. This decoupled architecture fully harnesses the semantic understanding capability of MLLMs and the fine‑grained perception advantage of the specialized grounding model, achieving leading performance on multiple benchmarks while avoiding the high costs of full fine‑tuning.
\end{itemize}

\section{Related Work}
\label{gen_inst}
\subsection{Autonomous GUI Navigation}
Early research explored GUI automation using rule-based engines or traditional machine learning~\cite{shi2017world,liu2018reinforcement,gurlearning,li2020mapping,lispotlight}. With the progress of large language models, LLM-driven agents have become mainstream, enhancing web interaction through in-context learning~\cite{zheng}, self-refinement~\cite{kim}, or specialized training for website interaction~\cite{gur2024real}. However, text-only LLMs cannot directly process visual interfaces and depend on GUI metadata or platform-specific customization~\cite{zheng,kim,li2020mapping,furuta2024multimodal,wen2024autodroid}. Recent vision-driven agents employing multimodal large language models show promise~\cite{cogagent,seeclick,shaw2023pixels,liu2024autoglm}, yet their performance remains constrained by the fine-grained perception limitations of MLLMs. Efforts to improve perception through specialized datasets face high construction costs and computational demands. This work proposes a decoupled framework that separates task planning from visual grounding, flexibly integrating an MLLM for semantic understanding with a specialized layout-aware model for precise localization, significantly improving accuracy while avoiding full fine-tuning.

\subsection{Multimodal Large Language Models}
The evolution of multimodal large language models (MLLMs)~\cite{team2024gemini,gpt4,qwenvl,qwen2.5VL,li2023blip,wang2023enabling,li2022blip, hou2026anchor} has unlocked new capabilities in cross-modal understanding. Early frameworks such as CLIP~\cite{CLIP} and ALIGN~\cite{align} established vision-language joint representation learning, while BLIP~\cite{li2022blip} introduced cross-attention mechanisms for visual question answering and image captioning. GPT-4V~\cite{gpt4} subsequently enabled visual in-context learning and inspired lightweight adaptation frameworks like LLaVA~\cite{liu2023visual} and MiniGPT-4~\cite{zhu2023minigpt}. Despite these advances, existing MLLMs face two critical challenges in GUI automation: insufficient fine-grained perception makes distinguishing visually similar elements difficult, and weak adaptability to dynamic layouts hinders the capture of spatial-semantic relationships of functional controls. Recent approaches address these limitations through domain adaptation---PIX2ACT++~\cite{shaw2023pixels} employs hierarchical screenshot encoding, while CogAgent~\cite{cogagent} enhances element localization via high-resolution visual tokenizers---yet both still depend on large-scale annotated data and fail to fully exploit inherent GUI design principles. This work proposes a task-deconstructed dual-stage localization framework that decouples GUI grounding into semantic parsing and visual grounding. By combining existing MLLMs' semantic understanding with a specially trained layout-aware localization model, we achieve precise element grounding without full fine-tuning, fully leveraging interface layout patterns to enhance cross-platform accuracy and robustness while reducing computational costs.

\subsection{Visual grounding}
In related work on visual grounding models\cite{glip,groundingdino,ren2024grounding}, approaches like GLIP\cite{glip} and Grounding DINO\cite{groundingdino} have significantly advanced open-vocabulary detection. GLIP reformulates object detection as a phrase grounding task, achieving language-aware visual representation learning, while Grounding DINO further enhances open-set object detection capability through its Transformer-based architecture. However, these methods still exhibit notable limitations in GUI automation scenarios: they are primarily designed for natural images, lacking sufficient understanding of the unique visual patterns and spatial layout characteristics specific to interface elements. We have developed an independently trained visual grounding model that substantially improves element localization accuracy and robustness by deeply exploring layout structure information within GUI interfaces. While extracting visual features, our model explicitly incorporates existing layout patterns in GUI interfaces, thereby overcoming the limitations of traditional methods in layout awareness for GUI contexts.

\section{Method}
\label{sec:method}

\noindent \textbf{Preliminary.}
GUI visual grounding translates natural language instructions into precise element coordinates on an interface screenshot $\mathcal{S} \in \mathbb{R}^{H \times W \times 3}$, outputting a bounding box $\mathbf{b} = (x_{\text{min}}, y_{\text{min}}, x_{\text{max}}, y_{\text{max}})$. Existing methods suffer from a fundamental trade-off: conventional grounding models lack semantic understanding, while end-to-end multimodal large language models (MLLMs) introduce coordinate hallucinations due to deficient fine-grained perception. We propose a decoupled intent-driven grounding framework that splits the task into instruction semantic elaboration and precise visual grounding. A training-free MLLM first produces structured visual descriptions from abstract instructions; then a Layout-Aware lightweight matching model performs accurate pixel-level grounding without full-scale fine-tuning. This separation allows the MLLM to focus on semantic reasoning while the specialized grounding model concentrates on fine-grained visual perception, effectively suppressing hallucinations. Details of each phase are given in Sections~\ref{sec:instruction-elaboration} and~\ref{sec:precise-localization}, with training procedures in Section~\ref{sec:training-details}.

\subsection{Instruction Elaboration}
\label{sec:instruction-elaboration}

The user's original instruction $U$ is typically expressed as an abstract functional description (e.g., ``click the login button'', ``navigate to the personal information page''), lacking specific visual cues that can be directly exploited for visual grounding. Inspired by the powerful cross-modal reasoning and commonsense knowledge of MLLMs, we introduce a frozen-parameter MLLM (e.g., Qwen-VL, GPT-4V) as the intent parser. In this stage, the MLLM receives the current screenshot $I$ and the user instruction $U$, and is guided via in-context prompt engineering to produce a structured visual description $Q$. This description precisely characterizes the visual attributes and spatial features of the target element. For example, ``click the login button'' is concretized as ``a blue rectangular button labeled `Login' located in the center of the page''. Since the MLLM does not undergo any gradient updates, this stage completely avoids the cost of large-scale model fine-tuning and the risk of catastrophic forgetting, while providing high-quality, semantically rich queries for downstream grounding.

\begin{figure*}[t]
    \centering
    \includegraphics[width=1.0\textwidth]{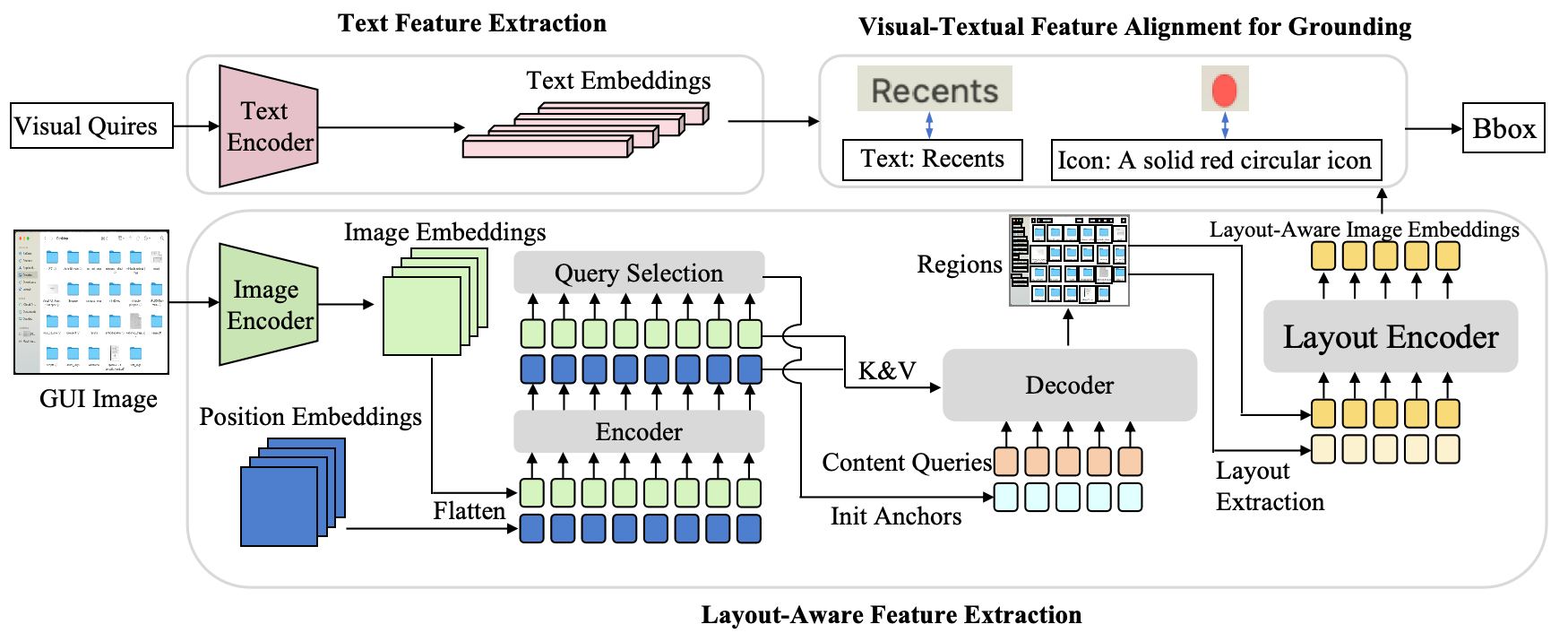}
    \caption{The overall architecture of the proposed layout-aware grounding model for GUI grounding, comprising three main stages: (1) text feature extraction, where visual queries are encoded into text embeddings to capture both textual and icon semantics of target elements; (2) layout-aware image feature extraction, where GUI image embeddings fused with position embeddings are encoded, followed by query selection to provide keys and values to a decoder that takes content queries and initial anchors as inputs to generate candidate regions, which are then fed into a layout encoder to produce layout-aware image embeddings; and (3) visual-textual feature alignment, where the text embeddings and layout-aware image embeddings are matched in a cross-modal manner to predict the bounding box of the target element.}
    \label{fig:model_architecture}
\end{figure*}

\subsection{Precise Localization}
\label{sec:precise-localization}

After obtaining the visual query $Q$, the subsequent pipeline is responsible for mapping it to a precise location in the image. In contrast to directly adopting an end-to-end learnable coordinate regression, we design a two-stage Layout-Aware grounding module that explicitly introduces a text/icon binary prior from the interface and geometric feature fusion, effectively suppressing coordinate hallucinations while avoiding full-scale fine-tuning.

\subsubsection{Layout-Prior Candidate Region Generation}
\label{sec:candidate-generation}

The first stage is handled by a GUI-domain-adapted DINO detector\cite{zhangdino}. We perform lightweight fine-tuning using only a small set of screenshots annotated with the two most basic layout types---``Text'' and ``Icon''. This design enables the detector to learn only to distinguish between text blocks and icon blocks during domain transfer, without involving any specific semantic categories; hence we refer to it as Layout-Aware candidate generation. The DINO Transformer decoder outputs an embedding $\mathbf{v}_i \in \mathbb{R}^{d}$ containing primary visual semantics for each predicted box. Meanwhile, we explicitly extract its normalized geometric attributes:
\begin{equation}
    \mathbf{g}_i = \left[ x_c,\ y_c,\ w,\ h,\ \frac{w}{h},\ w \cdot h \right],
\end{equation}
where $(x_c, y_c)$ are the relative coordinates of the region center, and $(w, h)$ are the relative width and height, all normalized to $[0, 1]$. This geometric vector reserves an interface for subsequent spatial enhancement.

\subsubsection{Hallucination-Resistant Cross-Modal Feature Matching}
\label{sec:feature-matching}

After obtaining candidate regions, we abandon the traditional learnable regression head and instead adopt a frozen-parameter feature matching strategy to produce the final coordinates. Specifically, we introduce the pre-trained vision-language model CLIP (ViT-B/32)\cite{radford2021learning} as the feature extractor. For each candidate region $r_i$, we crop the image and extract the visual feature $\mathbf{f}_i \in \mathbb{R}^{d_c}$ using the CLIP image encoder; meanwhile, the visual query $Q$ obtained from the first stage is encoded as a text feature $\mathbf{t} \in \mathbb{R}^{d_c}$. After $\ell_2$ normalization, we compute the cosine similarity:
\begin{equation}
    s_i = \frac{\mathbf{f}_i \cdot \mathbf{t}}{\|\mathbf{f}_i\| \|\mathbf{t}\|}.
\end{equation}
The candidate region with the highest similarity $i^* = \arg\max_i s_i$ is selected as the grounding result, and its bounding box is directly output as the predicted coordinates. This matching paradigm does not rely on learning any category-specific coordinate regression parameters, fundamentally eliminating the coordinate bias and hallucinations that may be introduced by full fine-tuning.

Simultaneously, to ensure the model's responsiveness to spatially sensitive queries (e.g., ``the return button in the upper left corner''), we incorporate an extremely lightweight geometric feature fusion module. For a candidate region $r_i$, its geometric vector $\mathbf{g}_i \in \mathbb{R}^6$ is first mapped via a learnable single-layer linear projection to a spatial offset of the same dimension as the CLIP visual feature:
\begin{equation}
    \mathbf{h}_i = \mathbf{W}_g \mathbf{g}_i + \mathbf{b}_g,
\end{equation}
where $\mathbf{W}_g \in \mathbb{R}^{d_c \times 6}$ and $\mathbf{b}_g \in \mathbb{R}^{d_c}$ are the only few learnable parameters. The offset is then additively fused into the original visual feature to obtain the layout-enhanced visual representation:
\begin{equation}
    \tilde{\mathbf{f}}_i = \mathbf{f}_i + \mathbf{h}_i.
\end{equation}
During matching, the enhanced $\tilde{\mathbf{f}}_i$ replaces the original $\mathbf{f}_i$ in the similarity computation. This design has very few learnable parameters, making training efficient. The additive fusion form allows the model to naturally learn an offset close to zero for spatially irrelevant queries, preserving CLIP's generalization ability; for spatially relevant queries, the geometric features provide positional compensation to the visual representation, steering the similarity scores toward regions that match the spatial expectation. The training of this module requires only a small number of samples with spatial annotations, while all other model parameters remain frozen.

\subsection{Training Details}
\label{sec:training-details}

\begin{figure}[t]
    \centering
    \includegraphics[width=1.0\linewidth]{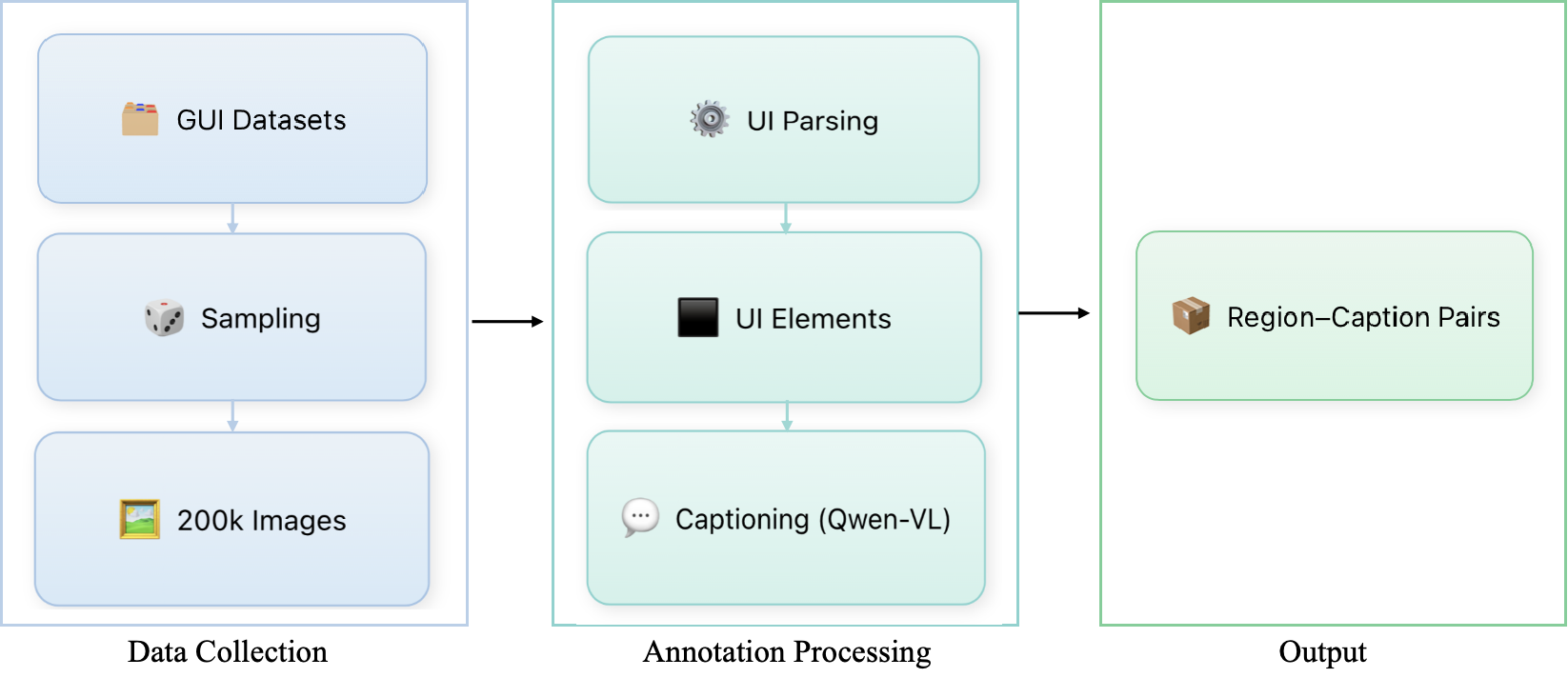}
    \caption{Training Dataset Construction Pipeline. We first sampled 200k images from common GUI datasets, then utilized parsing tools to extract icon and text regions from these images. Subsequently, we employed Qwen3-VL to generate image descriptions for these regions, ultimately producing region-caption pairs for model training.}
    \label{fig:data}
\end{figure}

\paragraph{Training Data Construction.}
We construct the training dataset following the pipeline illustrated in Figure~\ref{fig:data}. First, we sample approximately 200k screenshots from widely used GUI datasets. Using automated parsing tools\cite{omniparser}, we extract all textual and iconic regions from these images. Each extracted region is then fed into Qwen3-VL\cite{qwenvl} to generate a corresponding visual description, resulting in a collection of region-caption pairs.

\paragraph{Training Configuration.}
For the domain adaptation of the DINO detector (Stage~1), we use 200k GUI screenshots labeled only with ``Text'' and ``Icon'', split 9:1 for training and validation. The model employs a Swin-Tiny backbone initialized from COCO 12-epoch weights and is trained on 8 RTX 3090 GPUs with batch size 1 per GPU (effective 8) for 50 epochs. We use AdamW with base learning rate $1 \times 10^{-4}$, backbone learning rate $1 \times 10^{-5}$, weight decay $1 \times 10^{-4}$, and cosine annealing. Data augmentation consists solely of random scaling. All classification heads beyond the two layout classes are randomly initialized. In the matching and layout enhancement stage (Stage~2), we freeze CLIP ViT-B/32 and train only the geometric projection matrix $\mathbf{W}_g$ and bias $\mathbf{b}_g$, initialized from a standard normal distribution. We construct positive pairs from manually annotated spatial queries and target regions, with negatives sampled in-batch, and optimize using Adam with learning rate $1 \times 10^{-3}$ and batch size 256 for 20 epochs on the same GPU setup.

\paragraph{Loss Function.}
The domain adaptation stage for the DINO detector employs a multi-task detection loss following the standard DINO formulation:
\begin{equation}
    \mathcal{L}_{\text{det}} = \lambda_{\text{reg}} \mathcal{L}_{\text{reg}} + \lambda_{\text{giou}} \mathcal{L}_{\text{giou}} + \lambda_{\text{cls}} \mathcal{L}_{\text{cls}},
\end{equation}
where $\mathcal{L}_{\text{reg}}$ is the L1 regression loss for bounding box refinement, $\mathcal{L}_{\text{giou}}$ is the generalized IoU loss for spatial alignment, and $\mathcal{L}_{\text{cls}}$ is the focal loss for the binary classification between ``Text'' and ``Icon'' categories. Following standard practice, we set $\lambda_{\text{reg}}=5.0$, $\lambda_{\text{giou}}=2.0$, and $\lambda_{\text{cls}}=2.0$. The matching and layout enhancement stage is trained with a contrastive loss (InfoNCE). Formally, for a mini-batch of $B$ spatial query-region pairs, the loss is defined as:
\begin{equation}
    \mathcal{L}_{\text{align}} = -\frac{1}{B}\sum_{i=1}^{B}\log\frac{\exp(\text{sim}(\tilde{\mathbf{f}}_i, \mathbf{t}_i)/\tau)}{\sum_{j=1}^{B}\exp(\text{sim}(\tilde{\mathbf{f}}_j, \mathbf{t}_i)/\tau)},
\end{equation}
where $\text{sim}(\cdot,\cdot)$ denotes cosine similarity, $\tau$ is a temperature hyperparameter set to 0.07, and $\tilde{\mathbf{f}}_i$ and $\mathbf{t}_i$ are the geometry-enhanced visual feature and the text feature of the $i$-th description, respectively. The detection loss $\mathcal{L}_{\text{det}}$ is only applied during the domain adaptation stage, while the contrastive loss $\mathcal{L}_{\text{align}}$ is only applied during the matching and layout enhancement stage.

\section{Experiments}
In this section, we conducted experiments on both GUI grounding and GUI agent downstream tasks to validate the effectiveness of the proposed framework and the typography-aware grounding model. For the instruction elaboration phase, we employed Qwen2.5-VL-72B\cite{qwen2.5VL}.
\subsection{GUI Visual Grounding}
We comprehensively evaluate our method on the ScreenSpot\cite{seeclick} benchmark, a specialized dataset designed for visual grounding in professional application interfaces. This dataset contains 1,272 single-step interaction tasks collected from diverse professional software environments including productivity tools, creative applications, and enterprise systems across mobile (iOS/Android), desktop (macOS/Windows), and web platforms. For more rigorous evaluation, we extend our assessment to the ScreenSpot-Pro benchmark\cite{screenspot-pro}, a specialized dataset comprising 1,581 instruction-ground truth pairs collected from 23 professional applications across Windows, macOS, and Linux platforms. This dataset features meticulously annotated tasks with complex interface elements from specialized software environments including AutoCAD and Office suites. Both benchmarks encompass a wide spectrum of professional GUI components including textual elements in business contexts and domain-specific symbolic icons, providing comprehensive coverage of real-world professional usage scenarios.

\noindent \textbf{Compared Methods \& Evaluation.} Our comparative analysis encompasses several categories of models: advanced multimodal large language models including GPT-4o\cite{gpt4o}, Qwen2.5VL-72B\cite{qwen2.5VL}; specialized GUI understanding models such as SeeClick\cite{seeclick}, Fuyu\cite{fuyu}, CogAgent\cite{cogagent}, UGround\cite{uground}, and OS-Atlas-7B\cite{os-atlas}. Following established evaluation protocols for web agents, we employ click accuracy as our primary metric, defined as the percentage of test samples where the model's predicted coordinates fall within the ground truth element bounding box. This evaluation methodology maintains consistency with previous work like SeeClick, ensuring fair and comparable performance assessment across different approaches.

\noindent \textbf{Results.} As shown in Table \ref{tab:grounding1}, our method demonstrates superior performance compared to other models, achieving an average accuracy of 87.1\% across Mobile, Desktop, and Web platforms. Table \ref{tab:grounding2} further demonstrates its effectiveness with an average score of 41.3\% across six specialized domains. These results validate the efficacy of our proposed framework, where the task-decoupling design effectively leverages the complementary strengths of both MLLMs and dedicated grounding models to achieve robust localization performance.

\begin{table*}[htb]
\small
\begin{tabular}{lccccccc} 
\toprule
 &\multicolumn{2}{c}{\textbf{Mobile}} & \multicolumn{2}{c} 
{\textbf{Desktop}}& \multicolumn{2}{c}{\textbf{Web}} \\ 
\cmidrule(lr){2-3} \cmidrule(lr){4-5}  \cmidrule(lr){6-7} 
 \textbf{Model} & \textbf{Text} & \textbf{Icon/Widget} & \textbf{Text} & \textbf{Icon/Widget} & \textbf{Text} & \textbf{Icon/Widget} & \textbf{Average} \\ 
\midrule

Fuyu& 8.4& 6.6 & 6.2 & 2.9 & 6.5 & 3.4 &5.7\\
CogAgent& 8.4& 6.6 & 6.2 & 2.9 & 6.5 & 3.4 &5.7\\
GPT-4o& 25.6 & 24.5 & 14.4 & 19.3 & 9.6 & 6.3 &16.6  \\
Qwen2.5VL-72B & 29.0 & 26.2& 37.8 & 26.9 & 7.4 & 6.1 &22.2  \\
SeeClick& 78.0& 52.0 & 72.2 & 30.0 & 55.7 & 32.5 &53.4\\
Uground& 82.8	& 60.3& 82.5& 63.6	& 80.4 & 70.4 &73.3\\
\hline

Ours  & 95.2 & 81.4 & 94.4 & 78.9 & 92.7 & 80.1 & 87.1 \\
\bottomrule
\end{tabular}
\centering
\vspace{-3pt}
\caption{Results of different LVLMs on ScreenSpot.}

\label{tab:grounding1}
\end{table*}

\begin{table*}[htb]
\vspace{4pt}
\small
\begin{tabular}{lccccccc}
\toprule
\textbf{MODEL} & \textbf{Dev} & \textbf{Creative} & \textbf{CAD} & \textbf{Scientific} & \textbf{Office} & \textbf{OS} & \textbf{AVG} \\
\midrule

GPT-4o & 1.3 & 0.3 & 1.1 & 1.0 & 0.9 & 0.0 & 0.6 \\
SeeClick (7B) & 0.5 & 0.6 & 1.7 & 2.5 & 0.8 & 1.6 & 1.1 \\
MiniCPM-V (7B) & 2.8 & 1.3 & 3.6 & 5.4 & 2.8 & 2.3 & 3.0 \\
CogAgent (18B) & 7.3 & 5.3 & 6.2 & 13.4 & 9.2 & 3.0 & 7.7 \\
ShowUI (2B) & 10.1 & 4.2 & 4.4 & 10.9 & 12.9 & 6.6 & 7.7 \\
UGround-7B & 17.7 & 14.9 & 10.9 & 19.0 & 26.0 & 10.9 & 16.5 \\
OS-Atlas-7B & 21.3 & 16.4 & 9.9 & 25.3 & 26.2 & 17.4 & 18.9 \\
\hline
Ours  & 44.9 & 34.5 & 27.3 & 37.4 & 55.7 & 48.2 & 41.3 \\
\bottomrule
\end{tabular}
\centering
\vspace{-3pt}
\caption{Results of different LVLMs on ScreenSpot-Pro.}
\label{tab:grounding2}
\end{table*}
\vspace{-10pt}
\subsection{GUI Agents}
We evaluate the effectiveness of our method in both smartphone and web interaction environments using the Android in the Wild (AITW)\cite{rawles2023androidinthewild} and Mind2Web\cite{mind2web} benchmarks. The AITW dataset contains 30,000 natural language instructions and 715,000 multi-step operation trajectories, covering real-world mobile interaction scenarios including system settings adjustment, application switching, and form filling. The Mind2Web dataset focuses on web environments, comprising over 2,000 tasks from 137 real websites across 31 different domains, providing a diverse testing environment for building general web agents.

\noindent \textbf{Compared Methods \& Evaluation.}
We compare our approach with various state-of-the-art methods, including task-finetuned models such as Qwen-VL-7B\cite{qwenvl}, and SeeClick\cite{seeclick}, as well as  GPT-4V\cite{gpt4}, and Qwen2.5VL-72B\cite{qwen2.5VL} evaluated in a training-free setting. Our method maintains consistent output format with SeeClick, indicating target elements through coordinate outputs to ensure precise localization and interaction with GUI elements.
For evaluation metrics, we adopt the screen-wise action matching score as the primary metric in the AITW\cite{rawles2023androidinthewild} dataset, while computing Element Accuracy, Operation F1 score, and Step Success Rate for the Mind2Web dataset\cite{mind2web}. For vision-based methods, a prediction is considered correct when the predicted coordinates fall within the bounding box of the target element.

\noindent \textbf{Results.}
As shown in Table~\ref{tab:aitw}, our framework achieves an average improvement of 7.9\% over the baseline on the AITW dataset, with a 24.2\% gain in task success rate compared to using Qwen2.5VL-72B alone. On the more challenging Mind2Web dataset (Table~\ref{tab:mind2web}), our method achieves substantial gains in Element Accuracy and Step Success Rate across all task configurations. The Operation F1 score shows a narrower margin, which we attribute to our use of training-free MLLMs that trade task-specific optimization on action classification for stronger generalization.

\begin{table*}[htb]
\small
\begin{tabular}{lcccccc}
\toprule
\textbf{Model} & \textbf{General} & \textbf{Install} & \textbf{GoogleApps} & \textbf{Single} & \textbf{WebShopping} & \textbf{Overall} \\ 
\midrule

SeeClick & 54.0 & 66.4 & 54.9 & 63.5 & 57.6& 59.3   \\
QwenVL-7B\textsubscript{finetuned}  & 49.5& 59.9& 46.9 & 64.7 & 50.7 & 54.3\\
GPT-4V& 31.4& 32.1 & 29.1 & 39.7 & 22.5 & 31.0 \\

Qwen2.5VL-72B& 35.1& 50.1 & 51.6 & 41.2 & 39.2 & 43.4 \\
\hline

Ours& 61.5 & 71.4 & 60.6 & 72.9  & 63.5 &67.6\\
\bottomrule
\end{tabular}
\centering

\caption{Average scores of different methods on AITW. }
\label{tab:aitw}
\end{table*}

\begin{table*}[htb]
\centering
\footnotesize
\setlength{\tabcolsep}{3.5pt}
\begin{tabular}{lccccccccc}
\toprule
\textbf{Model} & \multicolumn{3}{c}{\textbf{Cross-Task}} & \multicolumn{3}{c} 
{\textbf{Cross-Website}}& \multicolumn{3}{c}{\textbf{Cross-Domain}} \\ 
\cmidrule(lr){2-4} \cmidrule(lr){5-7}  \cmidrule(lr){8-10} 
 & \textbf{Ele.Acc} & \textbf{Op.F1} & \textbf{Step SR} & \textbf{Ele.Acc} & \textbf{Op.F1} & \textbf{Step SR} & \textbf{Ele.Acc} & \textbf{Op.F1} & \textbf{Step SR} \\ 
\midrule
QwenVL-7B\textsubscript{finetuned}  & 15.9 & 86.7 & 13.3 & 13.2 & 83.5 & 9.2 &14.1&84.3&12.0\\ 
SeeClick& 28.3 & 87.0 & 25.5 & 21.4 & 80.6 & 16.4 &23.2&84.8&20.8\\ 
GPT-4V& 5.1 & 64.0&4.6  &5.2  & 61.0 & 4.2 & 5.2&62.7&4.6\\ 
CogAgent&22.4&53.0&17.6&18.4&42.2&13.4&20.6&42.0&15.5\\
Qwen2.5VL-72B& 27.7 &74.5&24.3&  25.6& 72.7 & 23.1 & 24.8 & 72.9&23.6\\ 
\hline

Ours &44.9& 75.8 & 41.2 &43.8  &75.4  & 39.3 & 40.1 &75.2&37.3\\ 
\bottomrule
\end{tabular}
\centering

\caption{Comparsion of methods on Mind2Web.}
\vspace{-5pt}
\label{tab:mind2web}
\end{table*}

\begin{figure}[t]
\begin{center}
\includegraphics[width=0.9\linewidth]{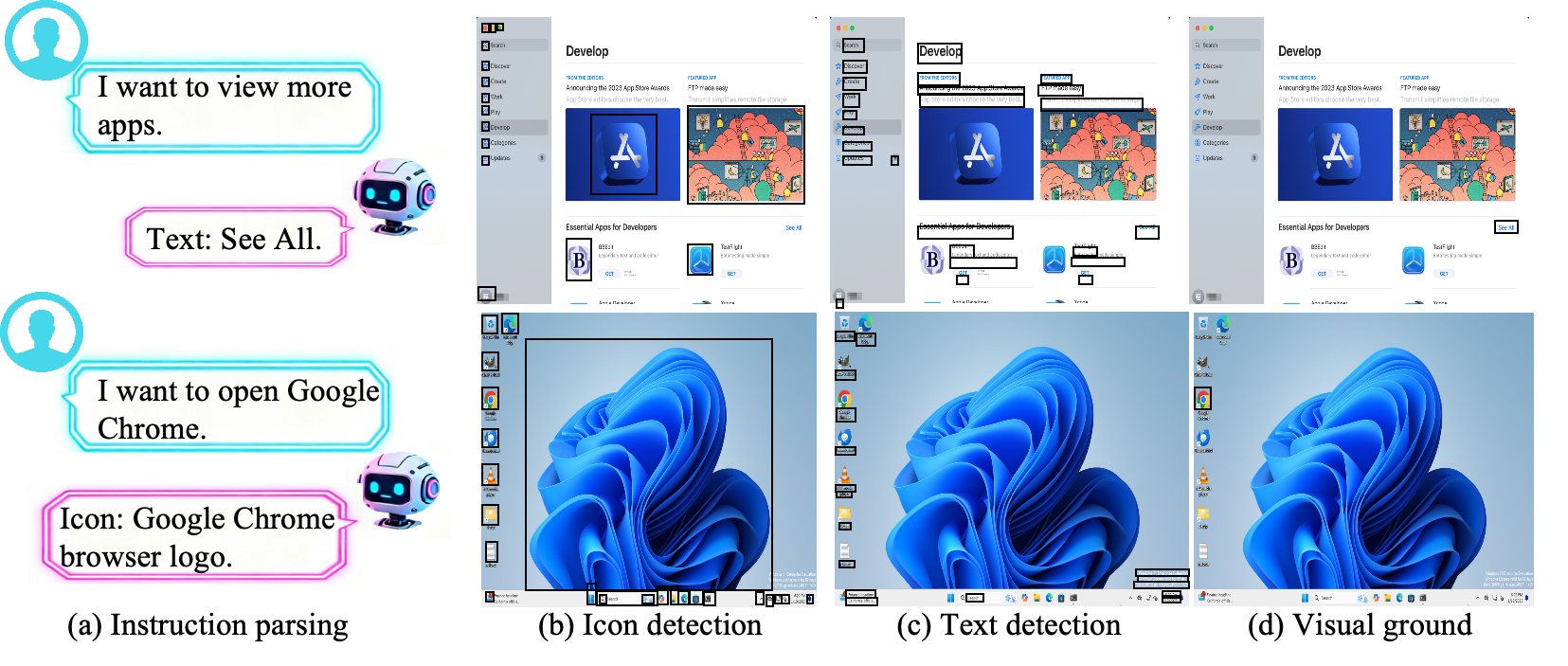}
\vspace{-0.5pt}
\end{center}
   \caption{(a) Instruction elaboration: abstract user commands are transformed into structured visual captions. (b)–(c) Icon detection and text recognition after pre-training. (d) Final localization via fine-tuning with visual captions.}
\label{fig:case}
\end{figure}

\subsection{Qualitative Results and Case Studies}

Figure~\ref{fig:case} illustrates the processing pipeline of our framework through two representative cases. Subfigure (a) shows abstract user commands transformed into structured visual captions by the frozen MLLM, capturing key visual attributes and spatial cues. Subfigures (b) and (c) demonstrate robust icon and text detection via layout‑prior candidate generation, effectively distinguishing text blocks from icon regions. Subfigure (d) presents the final precise localization results produced by regression‑free cross‑modal matching. Together, these visualizations validate the effectiveness of each design component and the seamless coordination across the entire grounding pipeline.

\section{Conclusion}
\label{sec:conclusion}

Our approach resolves the core dilemma in GUI grounding by decoupling instruction elaboration from precise localization. A frozen MLLM translates abstract instructions into structured visual descriptions; a dedicated Layout-Aware GUI Grounding Model then localizes targets via layout-prior candidate generation and regression-free cross-modal matching, inherently suppressing coordinate hallucinations without expensive fine-tuning. On ScreenSpot-Pro and Mind2Web, our method achieves over 20\% and 15\% improvement respectively over end-to-end state-of-the-art systems, with especially large gains on spatially sensitive queries---demonstrating a practical path toward hallucination-free GUI grounding.

This work contributes: (1) a Layout-Aware GUI Grounding Model that replaces coordinate regression with frozen cross-modal matching, preventing hallucinations while delivering precise localization; (2) a layout-prior injection and lightweight geometric fusion mechanism that acquires text understanding and spatial layout perception from minimal Text/Icon-level annotations; and (3) a decoupled intent-driven framework that separates instruction parsing from localization, harnessing MLLM semantics and specialized fine-grained perception without full fine-tuning.

\newpage
\small
\bibliographystyle{ieeenat_fullname}
\bibliography{main}




\end{document}